\documentclass[letterpaper,10pt]{article} 

\usepackage{opticameet3} 
\usepackage{caption}
\usepackage{booktabs}
\newcommand\authormark[1]{\textsuperscript{#1}}

\usepackage{amsmath,amssymb}
\usepackage[colorlinks=true,bookmarks=false,citecolor=blue,urlcolor=blue]{hyperref} 
\usepackage[font=footnotesize]{subcaption}

\begin{document}


\title{Experimental Demonstration of Online Learning-Based Concept Drift Adaptation for Failure Detection in Optical Networks}



\copyrightyear{2026}

\author{Yousuf Moiz Ali\authormark{1,*}, Jaroslaw E. Prilepsky\authormark{1}, Jo{\~a}o Pedro\authormark{2,3}, Antonio Napoli\authormark{4},\\ Sasipim Srivallapanondh\authormark{4}, Sergei K. Turitsyn\authormark{1}, and Pedro Freire\authormark{1,*}}

\address{\authormark{1}Aston Institute of Photonic Technologies, Aston University, B4 7ET Birmingham, UK\\
\authormark{2}Nokia, Optical Networks, 2790-078 Carnaxide, Portugal\\
\authormark{3} Instituto de Telecomunica\c{c}\~{o}es, Instituto Superior T\'{e}cnico, 1049-001 Lisboa, Portugal\\
\authormark{4}Nokia, Optical Networks, 81541 Munich, Germany}

\email{\authormark{*}y.moizali@aston.ac.uk, freiredp@aston.ac.uk} 


\vspace{-5mm}
\begin{abstract}
We present a novel online learning-based approach for concept drift adaptation in optical network failure detection, achieving up to a 70\% improvement in performance over conventional static models while maintaining low latency.
\textcopyright2026 The Author(s)
\end{abstract}

\vspace{-1.0mm}
\section{Introduction}
\vspace{-2mm}
Failure management is critical in optical networks to prevent service disruptions \cite{musumeci2025failure}. Machine learning (ML)-based methods have emerged as powerful tools for this task, thanks to their ability to enable partial or full system automation and reduce the likelihood of human error \cite{musumeci2019tutorial}. Several ML-based approaches have been proposed for failure detection and management in optical networks. However, these models are typically trained on historical data obtained from laboratory or field experiments and often struggle to adapt to non-stationary conditions, such as equipment aging or unexpected malfunctions. This lack of adaptability can lead to misclassifications, potentially degrading network performance and reliability.


One significant challenge is the occurrence of hard failures in optical networks, which are rare and difficult to simulate due to the limited data available from operators and vendors. Models trained primarily on soft failure data may struggle when encountering hard failures due to shift in the data distribution, a phenomenon known as concept drift (CD) \cite{lu2018learning}. While CD is inevitable in optical telemetry, ML models must adapt to it to sustain their performance over time.
Online learning, which trains models on sequential data received one event at a time \cite{hoi2021online}, is an efficient method for addressing the CD. To this end, we propose an online learning-based approach for adapting to CD for failure detection in optical networks. This approach enables continuous updates of the model in response to changing network conditions. To the best of our knowledge, this is the first work to study an online learning approach for failure detection in optical networks; the other online learning-related work in the field of optical networks focused on the prediction of the quality of transmission  \cite{wang2024lifelong}. Furthermore, this is the first work to consider methods for adapting to CD in optical network failure data. Our results indicate that online models provide an improvement of up to 70\% compared to static models in terms of rolling accuracy and increase the  Area Under the Curve (AUC) score to 0.75 while incurring only a minimal increase in latency.

\begin{figure*}[b!]
\vspace{-6mm}
  \centering
  \includegraphics[width=1.0\textwidth]{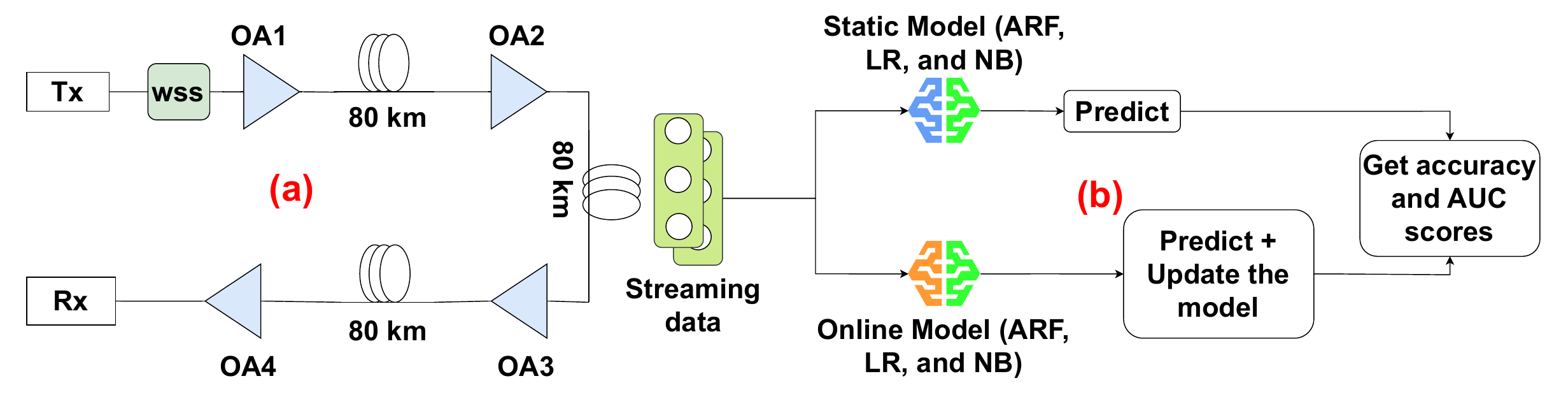}
  \captionsetup{font=footnotesize}
\caption{(a) Experimental testbed setup for failure detection where the Wavelength Selective Switch is used to introduce attenuation in OA1 to simulate normal and failure conditions. (b) Methodology used to test the static and online models. Assuming both the static and online models have been trained in a batch manner on SFD, the static model would only predict on the new sample, while the online model would predict and update the model with the new sample from the HFD.}
\vspace{-6mm}
\label{fig:methodology}
\end{figure*}



\vspace{-3mm}
\section{Methodology -- Online Training to Alleviate Concept Drift}
\vspace{-2mm}
To test our approach, we utilized the dataset generated by the authors in \cite{silva2022learning}. The testbed used to generate this dataset is shown in Fig.~\ref{fig:methodology}a. This dataset was divided into two parts: the Hard Failure Dataset (HFD) and the Soft Failure Dataset (SFD). The features generated for both datasets contain the Timestamp, Type of device, ID of the device, Bit Error Rate (BER), and Optical Signal-to-Noise Ratio (OSNR) for the Transmitter (Tx) and Receiver (Rx). In this work, only the BER and OSNR of the Tx and Rx serve as the final features for training the model. To replicate real-world optical telemetry, we merged the two datasets, where the SFD to HFD transition was treated as a temporal drift event to mimic the real-world evolution of network health conditions. We applied the SFD to train both the static and online models in a batch manner. The HFD was used as streaming data, where every sample from the data was utilized to predict on the static model and to predict and update the online model, as shown in Fig.~\ref{fig:methodology}b, thereby replicating online learning. The reasoning behind this is that hard failures are very rare in a real optical network; thus, most static models would be trained on soft failures and struggle when a hard failure occurs at any point in time. This approach provides a clear understanding of how static models deteriorate in performance when the data distribution undergoes a transition from soft failure to hard failure, and also highlights the benefits of online learning in adapting to CD. We currently only consider online supervised learning, where the online model already has access to the labels. Note that, while we only examined a soft-to-hard failure CD scenario in this paper, the results presented here are also applicable to a soft-to-soft failure drift scenario.

\begin{figure*}[t!]
\vspace{-6mm}
  \centering
  \includegraphics[width=\textwidth]{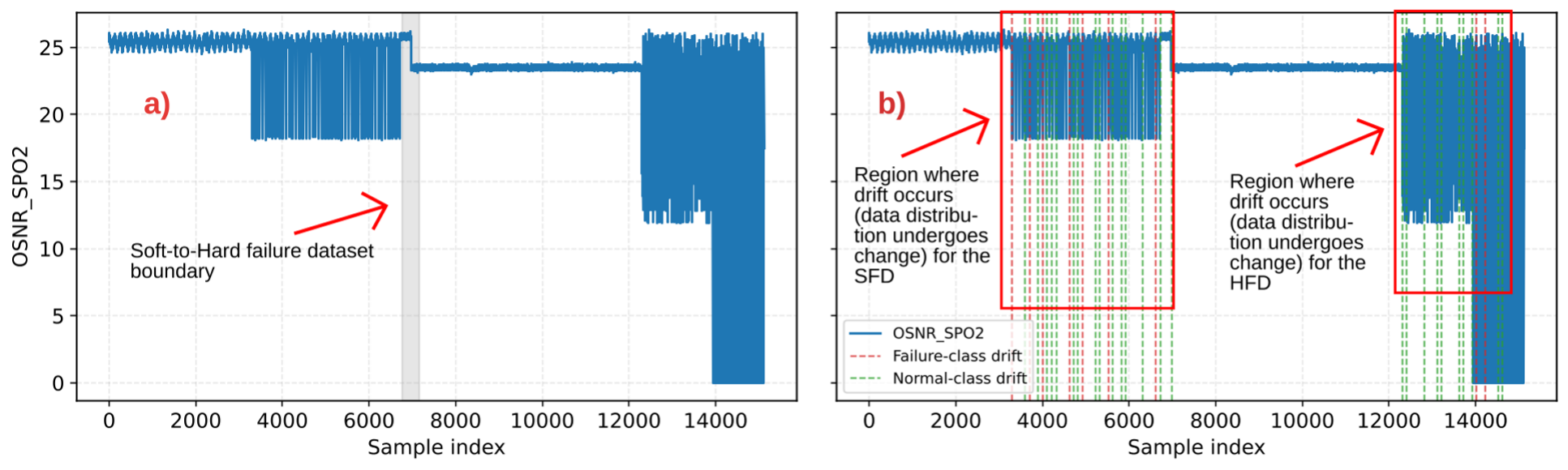}
  \captionsetup{font=footnotesize}
\caption{a) Data distribution for \textit{OSNR\_SPO2} feature for the combined dataset of soft and hard failures, showing the drop in OSNR is higher for HFD than SFD after a period of normal instances. b) Plot showing specific areas where drift has occurred in the \textit{OSNR\_SPO2} feature with green lines showing drift in the normal class (without failure) and red lines showing drift in the failure class.}
\vspace{-6mm}
\label{fig:CD}
\end{figure*}

The next step of our approach was to detect any CD in the dataset. For this, we utilized the Page-Hinkley Test (PHT). PHT is a combination of the work in \cite{page1954continuous} and \cite{hinkley1970inference}. PHT compares the cumulative sum of the data points with its running minimum, signaling drift when the difference exceeds a pre-defined threshold. Fig.~\ref{fig:CD}a shows the data distribution for the \textit{OSNR\_SPO2} (OSNR at the receiver) feature for the combined dataset. Since the feature variable \textit{OSNR\_SPO2} exhibits the highest correlation with the target variable, we only considered drift within this feature due to its maximal influence on the output. We can see in Fig.~\ref{fig:CD}a that the drop in OSNR is higher for HFD than SFD, signaling a change in the data distribution. Fig.~\ref{fig:CD}b shows the identified drifts in the dataset; every failure-class drift in the SFD is preceded by some normal-class drifts; however, in the HFD, failure-class drifts occur suddenly without any prior warning. 

We chose the Adaptive Random Forest (ARF), a variant of the Random Forest algorithm tuned for online learning \cite{gomes2017adaptive}, Logistic Regression (LR), and the Naive Bayes (NB) algorithm as the ML models to be trained on this dataset. These algorithms provide a representative spectrum of model complexities, from linear to ensemble-based learners, to benchmark online learning robustness. All the experiments were performed on an Intel i5 13th Gen CPU with 16 GB of RAM, and no GPU acceleration was used. The parameter values of the models are listed in the last column of Tab.~\ref{tab:latency}. The metrics used for the evaluation are the rolling accuracy and the AUC score, where accuracy and AUC are monitored and recorded over windows of 500 data points.

\vspace{-3mm}
\section{Results and Discussion}
\begin{figure*}[t!]
\vspace{-6mm}
  \centering
  \includegraphics[width=\textwidth]{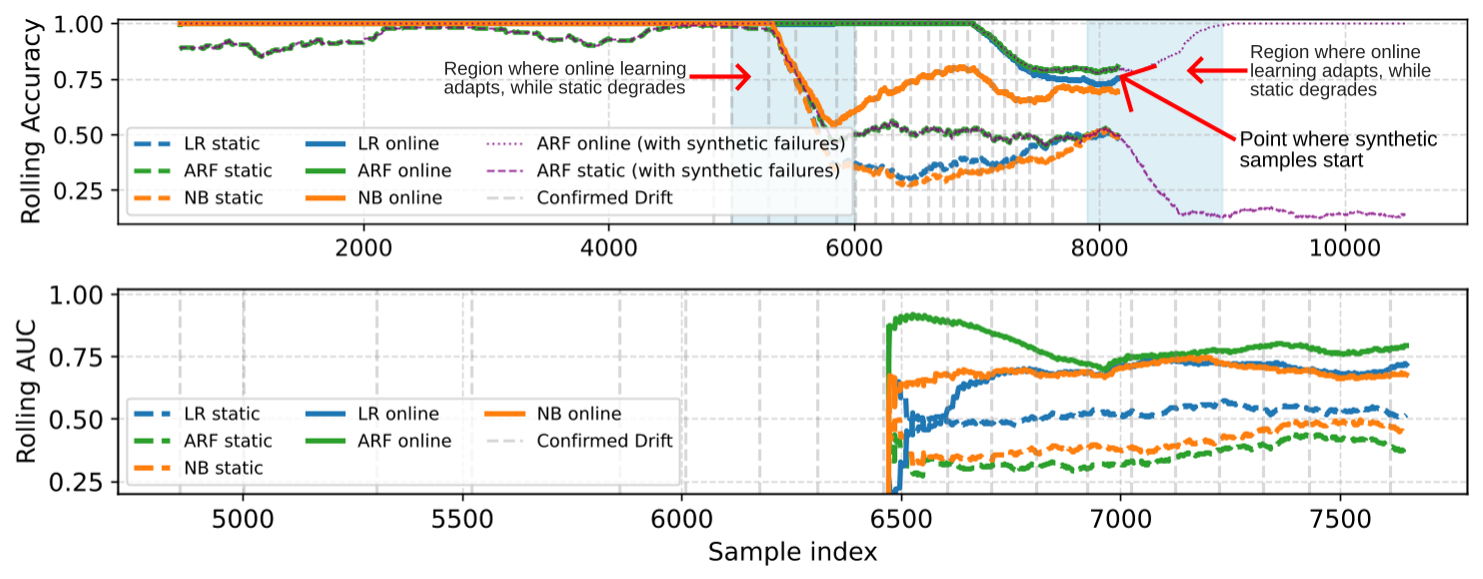}
  \captionsetup{font=footnotesize}
  \vspace{-8mm}
\caption{Rolling accuracy and AUC plots on the HFD. The shaded regions represent those areas where the online models maintain or improve performance during drift regions, whereas static models degrade in performance. The arrow represents the point after which synthetic failure samples were available to show that if more failure samples are added, the online model regains accuracy to 100\% after the earlier dip, but the static model further degrades.}
\vspace{-8mm}
\label{fig:rolling_metric}
\end{figure*}

\vspace{-2mm}
Fig.~\ref{fig:rolling_metric} shows the rolling accuracy and AUC plots for all three ML models. The plot shows that the online models for ARF, LR, and NB provide a higher level of performance than the static models in terms of both rolling accuracy and rolling AUC score. Examining the rolling accuracy plot, the online LR model achieves the most significant improvement, up to 70\%, compared to the static model, while the online ARF yields an improvement of up to 55\% relative to the static model. If we compare only the online models, then ARF provides the most sustained performance compared to LR and NB, primarily due to its ability to learn complex patterns. It can clearly be seen from the accuracy plot that as soon as the static models encounter drift, the performance drops and does not recover to the same value as before. In contrast, the online models for ARF and LR only display a dip in performance during the hardest drift regions, regaining performance as the model adapts and observes more samples from the failure class. Random oversampling was used to increase the HFD to create synthetic failure samples. The online ARF model rapidly adapts to these new samples, regaining 100\% accuracy, whereas the static ARF model exhibits performance degradation due to its limited generalization to unseen failure patterns, as shown in Fig.~\ref{fig:rolling_metric}.

Focusing on the rolling AUC plot, we observe a trend similar to that of the rolling accuracy plot. The static models' AUC scores hovered around the 0.5 mark, indicating that the models were guessing randomly and unable to differentiate between normal and failure samples. The online models performed significantly better than the static models, with their AUC scores experiencing dips when data drift occurs, but then showing an upward trend and stabilizing around the 0.75 mark. These results clearly demonstrate the advantage of online learning in the face of evolving network dynamics and CD in the data. Furthermore, the results also indicate that online learning is model-agnostic, as it consistently outperforms static models, regardless of the ML model under consideration. It is worth noting that this is just one scenario of how CD can impact optical networks. There may be other instances, such as the CD within different soft failures, but the approach should also be applicable to those scenarios.

Lastly, we consider the latency increase when using online learning. Tab.~\ref{tab:latency} shows the median per-event latency for the static and the online models for 100 individual trials. For static models, we calculated the median time to predict on a single sample. For online models, we calculated the median time to predict and update the model on that sample. As shown in the table, online models result in an increased latency, with the ARF being the most computationally intensive. However, with the online overhead remaining $\sim$ 1~ms and below, the performance gain can fully justify this increase in latency.

\vspace{-3mm}
\section{Conclusion}
\vspace{-2mm}
In this work, we presented a novel online learning-based approach for adapting to concept drift in optical network failure detection. Our results show that online models (namely, ARF, LR, and NB) achieved up to a 70\% improvement in rolling accuracy compared to static models. Rolling AUC scores exhibited a similar trend, with online approaches consistently outperforming their static counterparts. Furthermore, the proposed online models demonstrated effective adaptation to evolving data distributions while maintaining low computational complexity, incurring a latency of less than 1~ms.

\vspace{-2mm}

\begin{table}[htb]
\centering
\renewcommand{\arraystretch}{1.15} 
\setlength{\tabcolsep}{5pt} 
\label{tab:latency}
\footnotesize
\begin{tabular}{@{}lccc p{6.5cm}@{}}
\toprule
\textbf{Model} & \textbf{Static (ms)} & \textbf{Online (ms)} & \textbf{Overhead (ms)} & \textbf{Parameter Values} \\ 
\midrule
LR  & 0.0015 & 0.0094 & 0.0078 & Optimizer: Stochastic Gradient Descent; Loss: Binary log loss; Learning rate: 0.01 \\[2pt]
NB  & 0.0769 & 0.0830 & 0.00635 & Minimum variance allowed: $1\times10^{-10}$ \\[2pt]
ARF & 0.136  & 0.404  & 0.258 & Number of trees: 10; Max features: 2 (features considered when splitting); Samples before split: 50 \\
\bottomrule
\end{tabular}
\captionsetup{font=footnotesize}
\vspace{-2mm}
\caption{Median per-event latency for static and online models over 100 trials, including online overhead. The final column lists the parameter values for each model.}
\label{tab:latency}
\end{table}

\vspace{-6mm}
\footnotesize
\textbf{Acknowledgments} This work has received funding from the European Commission MSCA-DN NESTOR project (G.A. 101119983). SKT acknowledges EPSRC project TRANSNET (EP/R035342/1). Antonio Napoli and Sasipim Srivallapanondh acknowledge the EU HORIZON SENSEI project (GA No. 101189545).

\end{document}